\documentclass[10pt,journal,compsoc]{IEEEtran}

\ifCLASSOPTIONcompsoc
  \usepackage[nocompress]{cite}
\else
  \usepackage{cite}
\fi

\ifCLASSINFOpdf
\else
\fi
\usepackage{color}
\usepackage{algorithm}
\usepackage{algorithmic}
\usepackage{caption}
\usepackage{xspace}
\usepackage{graphicx}
\usepackage{multirow}
\usepackage{ragged2e}
\usepackage{amsmath,amssymb}
\usepackage{makecell}
\usepackage{colortbl}
\usepackage{xcolor}
\usepackage{array}   
\usepackage{pifont}

\newcommand{\xmark}{\ding{55}}
\newcommand{\model}{TDMR\xspace}

\def\ie{\mbox{\textit{i.e.}, }}
\def\eg{\mbox{\textit{e.g.}, }}

\def\CE{\mathrm{CE}}
\def\KL{\mathrm{KL}}

\def\mD{{\mathcal D}}

\def\mH{{\mathcal H}}

\def\mL{{\mathcal L}}

\def\mX{{\mathcal X}}

\DeclareMathAlphabet\mathbfcal{OMS}{cmsy}{b}{n}

\def\0{{\bf 0}}
\usepackage{ dsfont }
\def\1{\mathds{1}}

\def\bx{{\bf x}}

\def\mmE{{\mathbb E}}

\def\bx{{\bf x}}

\usepackage{mathrsfs}

\usepackage{ntheorem}
\newtheorem{thm}{Theorem}

\begin{document}

\title{Towards Efficient Task-Driven Model Reprogramming with Foundation Models\xspace}

\author{Shoukai~Xu,
        Jiangchao~Yao, 
        Ran~Luo,
        Shuhai~Zhang,
        Zihao~Lian,
        Mingkui~Tan,
        Bo~Han,
        and~Yaowei~Wang
\IEEEcompsocitemizethanks{
\IEEEcompsocthanksitem{Shoukai Xu, Ran Luo, Shuhai Zhang, Zihao Lian and Mingkui Tan are with the School of Software Engineering, South China University of Technology, Guangzhou, China, 510006 (e-mail: sexsk@mail.scut.edu.cn; 202121047019@mail.scut.edu.cn; mszhangshuhai@mail.scut.edu.cn; 202121046792@mail.scut.edu.cn; mingkuitan@scut.edu.cn).\\
Jiangchao Yao is with Cooperative Medianet Innovation Center, Shanghai Jiao Tong University, Shanghai, China (e-mail: sunarker@sjtu.edu.cn).\\
Bo Han is with Hong Kong Baptist University, HongKong, China (e-mail: bhanml@comp.hkbu.edu.hk).\\
Yaowei Wang is with PengCheng Laboratory, Shenzhen, China (e-mail: wangyw@pcl.ac.cn).
}
\IEEEcompsocthanksitem{Shoukai Xu and Jiangchao~Yao are equal contributors.}
\IEEEcompsocthanksitem{Yaowei~Wang, Bo~Han and Mingkui~Tan are corresponding authors.}
}}

\markboth{Journal of \LaTeX\ Class Files,~Vol.~14, No.~8, August~2015}%
{Shell \MakeLowercase{\textit{et al.}}: Towards Efficient Task-Driven Model Reprogramming with Foundation Models\xspace}

\IEEEtitleabstractindextext{
\begin{abstract}

Vision foundation models exhibit impressive power, benefiting from the extremely large model capacity and broad training data. However, in practice, downstream scenarios may only support a small model due to the limited computational resources or efficiency considerations. 
Moreover, the data used for pretraining foundation models are usually invisible and very different from the target data of downstream tasks.
This brings a critical challenge for the real-world application of foundation models: one has to transfer the knowledge of a foundation model to the downstream task that has a quite different architecture with only downstream target data.  
Existing transfer learning or knowledge distillation methods depend on either the same model structure or finetuning of the foundation model.
Thus, naively introducing these methods can be either infeasible or very inefficient.
How to leverage the knowledge from the foundation model to boost the small model has not been well studied.
To address this, we propose a Task-Driven Model Reprogramming (\model) framework.
Specifically, we reprogram the foundation model to project the knowledge into a proxy space, which alleviates the adverse effect of task mismatch and domain inconsistency. 
In this stage, we maintain the foundation model as a powerful feature extractor frozen.
Then, we reprogram the target model via progressive distillation from the proxy space to efficiently learn the knowledge from the reprogrammed foundation model.
\model is compatible with different pre-trained model types (CNN, transformer or their mix) and limited target data, and promotes the wide applications of vision foundation models to downstream tasks in a cost-effective manner. 
Extensive experiments on different downstream classification tasks and target model structures demonstrate the effectiveness of our methods with both CNNs and transformer foundation models. For example, on CUB-200, \model improves the accuracy of MobileNetV2 from 62.90\% to 72.60\% using the ResNet-50 as a teacher and to 76.04\% using the Swin transformer as a teacher.
\end{abstract}

\begin{IEEEkeywords}
Foundation model, model reprogramming, knowledge distillation, transfer learning, downstream task.
\end{IEEEkeywords}}

\maketitle

\IEEEdisplaynontitleabstractindextext

\IEEEpeerreviewmaketitle

\section{Introduction}\label{Introduction}
\IEEEPARstart{P}{retraining} methods powered by the continual evolution of neural architectures and broad data achieve great success in extensive vision tasks~\cite{han2022survey,bommasani2021opportunities}. The resulting foundation models, also called big models, thus have been recognized as a general solution in industrial and academic areas~\cite{touvron2022resmlp, yuan2022volo, Devlin2019bert, brown2020language, radford2021learning, liu2021swin, Bao2022beit}.
Despite impressive performance on visual benchmarks, it is still limited in utility due to the expensive computational cost~\cite{li2022ds,mou2022transcl}.
Especially the constraints of computing power and model size
in real-world downstream scenarios all make it impossible to use  the foundation model directly.
Due to the inherent conditions, only target scene data and suitable target models can be used, and how to broaden its scope of applications in real-world scenarios is still an open problem.

\begin{figure*}[t]
	\centering
	\includegraphics[width=0.95\linewidth]{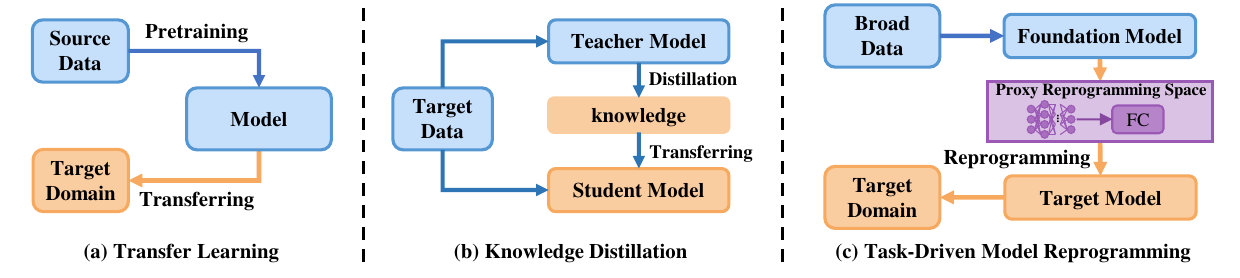}
 \caption{Comparisons of different learning paradigms. (a) Transfer learning targets at reusing a model pretrained on source data to target domain. (b) Knowledge distillation seeks to exploit the knowledge from the teacher to guide the student on the same data. (c) Different from the both, our task-driven reprogramming aims to employ a foundation model learned on the broad data to improve a model for target domain.
Note that, this is also different from the ordinary model reprogramming that only considers to repurpose the foundation model. 
}
	\label{bottleneck}
\end{figure*}

To enable the utility of foundation models in constrained scenarios, it is imperative to refine the foundation model into a target model that is capable of effectively addressing specific downstream tasks.
However, there are significant gaps in model structures and data domains between the foundation model and the downstream target model.
Concretely, the foundation model and target model for downstream tasks usually have significant differences in architecture.
Moreover, the broad data used for pretraining foundation models are generally collected from wild domains, but the target data of downstream tasks is often within a dedicated domain. 
Hence, applying the vision foundation model to downstream tasks should not only complete the domain transfer but also include distillation from large models to small models.

Most existing methods can only handle a certain aforementioned gap while ignoring the other aspects.  
The popular transfer learning methods~\cite{ramirez2023learning,liang2021source,dong2021and,kang2020contrastive,long2018transferable, lin2022prototype} aim to reuse the model from the old domain to the new one, which does not consider the cross-architecture problem, as shown in Figure~\ref{bottleneck} (a). 
Ordinary knowledge distillation methods~\cite{hinton2015distilling,ye2022dynamic,wang2021knowledge,zhuang2021effective,zhang2021self,tian2019contrastive,zhang2022minivit,liu2022cross} use the supervision information provided by large models to train small models for better performance, which builds upon the same data and the same task, as shown in Figure~\ref{bottleneck} (b). They may face challenges when addressing divergences in data domains.
A recent paradigm, model reprogramming~\cite{chen2022model,jia2022vpt,chen2022adaptformer}, shows potential to transfer scenarios. However, the normal model reprogramming methods are still limited in repurposing the pretrained foundation model without considering the downstream model.
Therefore, the above methods cannot efficiently solve the targeting challenges in the application of foundation models.

In this work, we propose an efficient Task-Driven Model Reprogramming (\model) framework (as shown in Figure~\ref{bottleneck} (c)) to jointly consider the aforementioned challenges when applying foundation models to downstream tasks. 
Different from methods based on either pure knowledge distillation (\textbf{KD}) or pure model reprogramming (\textbf{MR}), our TDMR has the merits of both paradigms.
TDMR applies foundation models to downstream scenarios across inconsistent data domains and diverse architectures.
Moreover, different from the naive combination of knowledge distillation and model reprogramming paradigms, the most distinct point is that our TDMR is well-designed for the application of the foundation model on various downstream tasks.
First, we design a universal framework that enables the foundation model to be efficiently customized to a wide range of downstream tasks. Second, we devise a general way to promote the effective transfer of implicit knowledge from the pre-trained foundation model to the target model, transcending both data domains and model structures.

Specifically, we first propose to reprogram the foundation model (the teacher) by constructing a proxy space that facilitates the cross-domain interchange of knowledge for the foundation model and the target model (the student).
Depending on the downstream scenario, a projector followed by a task-aligned classifier is applied together to form the proxy space, which helps alleviate the mismatch in various downstream tasks.
Second, our distillation in the reprogramming space is progressive, namely, first transfer the classifier to tune the feature backbone, and then jointly optimize the classifier and the feature backbone, which guarantees that the knowledge in the reprogramming space is mildly but more efficiently transferred.
Third, different from previous methods based on KD or model reprogramming that requires the specific design adaptation according to pre-trained model types (CNN, transformer or their mix), our TDMR is totally compatible.
This will give full play to the advantages of vision foundation models, thereby promoting more applications to the downstream scenarios.
In a nutshell, our contributions can be summarized as the following three points: 

\begin{enumerate}
\item We establish a new task-driven model reprogramming (TDMR) framework to promote the wide applications of foundation models in downstream tasks, which supports efficient knowledge transfer among inconsistent domains and diverse architectures. 
Our TDMR is compatible with different pre-trained model types (CNN, transformer or their mix) and limited target data, which fully considers the actual situation in the application of foundation models.

\item 
\model has dedicated designs for applying foundation models to downstream tasks.
With the purpose of improving the efficiency of knowledge transfer, \model constructs a proxy reprogramming space to align the task prototype and reduce domain inconsistency. Then in order to ensure the knowledge in the reprogramming space is mildly but more efficiently transferred, \model reprograms the target model via the progressive distillation.

\item We theoretically analyse how the domain gap affects the performance of the transfer and utilize \textit{maximum mean discrepancy (MMD)} between two domains as auxiliary supervision to improve performance if some broad data is available.
Moreover, if some broad data is available, we can use MMD between the domains of broad data and downstream target data to further make the target models achieve better performance.

\item 
Extensive experiments on various downstream tasks and diverse model structures show the superior performance of \model, and a range of ablation studies show the rationality of our design. We also empirically verify the merit of \model in reducing the annotation of the downstream task under semi-supervised learning.
\end{enumerate}

\section{Related Work}\label{relatedwork}
We first review the development of vision foundation models. Then, we discuss relevant topics that aim to apply foundation models to downstream tasks, \ie{knowledge distillation, transfer learning and model reprogramming}.

\textbf{Vision Foundation Models.}
Recently, a range of works has explored the powerful architectures for foundation models~\cite{dosovitskiy2020image,wu2022p2t,liu2021swin,qian2022makes,peng2023conformer,sun2022patch}.
Alexey et al.~\cite{dosovitskiy2020image} proposed Vision Transformer structures (ViT) by dividing images into patches and casting them as a sequence of tokens as input, which has quadratic computation complexity \textit{w.r.t.,} image size. 
Ze et al.~\cite{liu2021swin} proposed a Swin Transformer that constructs progressive feature maps and computes self-attention locally within non-overlapping windows, reducing to the linear computational complexity \textit{w.r.t.} image size. 
To overcome the drawback of the transformer that mainly focuses on the long-range dependencies, Peng et al.~\cite{peng2023conformer} proposed Conformer to fuse CNN-based local features with transformer-based global representations. Due to the impressive performance in computer vision tasks, models based on ViT are further developed like Scaling Vision Transformer~\cite{chen2022scaling}, Pyramid Vision Transformer (PVT)~\cite{wang2021pyramid}, Vision Outlooker (VOLO)~\cite{yuan2022volo} and Sparse ViT~\cite{riquelme2021scaling}.

\textbf{Knowledge Distillation.}
\label{relatedwork:KD}
It is an effective model compression technique that transfers knowledge from teacher to student, which can be summarized from the perspectives of logit distillation~\cite{hinton2015distilling,huang2022knowledge,zhao2022decoupled,chen2022knowledge}, representation distillation~\cite{wang2021distilling,tian2019contrastive,ji2021show,chen2021reviewkd,yang2022masked} and relation distillation~\cite{ye2022generalized,zhang2021self,yim2017gift,park2019relational,passalis2020heterogeneous}. 
Hinton et al.~\cite{hinton2015distilling} proposes knowledge distillation and uses Kullback-Leibler divergence to align student's logits with teacher's. Then many works focus on this technique. 
Logit distillation methods~\cite{hinton2015distilling,huang2022knowledge,zhao2022decoupled} utilize the logits of teacher as soft labels to supervise student training. Feature distillation methods~\cite{romero2014fitnets,zagoruyko2016paying,tian2019contrastive,ji2021show} choose the features of intermediate layers as the media transferring the dark knowledge. 
Moreover, the relationships of intermediate layers' features are also considered in \cite{yim2017gift,park2019relational,passalis2020heterogeneous}. 
It is inspiring that these methods provide elegant ways to leverage cumbersome models and obtain considerable performance gains. 
Due to the different structures of CNN and Transformer, existing knowledge distillation methods focusing on CNN cannot be directly applied to Transformer.
There are some works~\cite{touvron2021training, zhang2022minivit, wu2022tinyvit, liu2022cross} focused on knowledge distillation for transformer. Touvron et al.~\cite{touvron2021training} propose a token-based strategy called DeiT, which uses a distillation token to the student Transformer to learn the hard label from the teacher model. Liu et al.~\cite{liu2022cross} propose a novel cross-architecture knowledge distillation method to bridge the large gap between the transformer and CNN, which uses several carefully design projectors to match the teacher (the transformer teacher model) and the student (the CNN model) in the same feature space. To transfer knowledge from large-scale ViT models to student transformer, MiniViT~\cite{zhang2022minivit} uses weight distillation over self-attention and multiplexes the weights of consecutive transformer blocks while imposing a transformation on the weights to increase diversity. To save memory and computation overhead, Wu et al.~\cite{wu2022tinyvit} store the sparse soft labels of the large teacher model in disk and the small student transformer models can learn directly from these labels.
Recently, Ye et al.~\cite{ye2020distilling} proposed cross-task knowledge distillation, which leverages a relation-based distillation method to handle the non-overlapping label spaces in the homogeneous domain. 
However, it is difficult for existing distillation methods to use the pretrained teacher model on cross-domain data.
Different from them, we target to solve the heterogeneous cross-domain problem.

\begin{table}[]
\centering
\caption{Comparison of different methods when applying the foundation model to downstream tasks from the perspective of \emph{domain} and \emph{structure}.}
\setlength{\tabcolsep}{1.45mm}
\renewcommand\arraystretch{1.3}
{
\begin{tabular}{c|cc}
\hline
 Methods & Different Domains & Different Structures \\ \hline
 Vanilla KD &         \xmark              & \checkmark \\
 Feature KD &         Weak                & \checkmark \\
 Transfer Learning &    \checkmark        & \xmark  \\
 Model Reprogramming &  \checkmark        & \xmark \\
 TDMR &                 \checkmark        & \checkmark \\ \hline
\end{tabular}
}
\label{tab:relatedworks}
\end{table}

\textbf{Transfer Learning.} Transfer learning aims to improve the performance on the target domain via discovering and transferring latent knowledge from source data~\cite{kouw2019review,ramirez2023learning,liang2021source,dong2021and}, which has wide applicability.
To this end, Unsupervised Domain Adaptation (UDA) methods~\cite{ganin2016domain, li2021synthetic} seek to conduct domain alignment between a label-rich source domain and an unlabeled target domain. 
Unlike traditional UDA methods, source-free UDA methods~\cite{liang2020we, Li2020ModelAU} mitigate domain discrepancy between an unlabeled target domain and a source model instead of accessing source data. To transfer knowledge from the source data, various finetune methods~\cite{oquab2014learning, abnar2022exploring, yamada2022does} add task-specific layers to the pretrained model to suit the target task. Recently, Jia et al.~\cite{jia2022vpt} proposed Visual Prompt Tuning (VPT) to save computational costs, which introduces several learnable prompts in patch embedding phrases without finetuning.
However, the above methods could alleviate discrepancies in domains and tasks while all of them are unable to address the issue of knowledge transfer between diverse architectures.

\textbf{Model Reprogramming.} Although finetuning is a straightforward way to leverage a pretrained model, the enormous computational and time costs of vision foundation models make finetuning impractical in downstream computing-prohibitive applications.
Therefore, reprogramming foundation models with only a few learnable parameters becomes a popular choice recently~\cite{chen2022model}. To this intuition,  
Elsayed et al.~\cite{elsayed2018adversarial} proposed adversarial reprogramming, which just maps the imageNet labels to task labels, to confuse the ImageNet classifier to be the CIFAR-10 or MNIST classifier. Dinh et al.~\cite{dinh2022improved} revealed that with scarce labeled data, input reprogramming is better than training from scratch and finetuning. Tsai et al.~\cite{tsai2020transfer} proposed black-box adversarial reprogramming, which transforms the input and adds the label mapping machinist to the output. Note that, model reprogramming only trains the inserted input transformation and output mapping layers~\cite{neekhara2022cross, neekhara2018adversarial, kloberdanz2021improved, yang2021voice2series}, while our~\model is a more generalized extension, which reprograms the foundation model to the target model for the purpose of knowledge transfer.

\begin{figure*}[t]
  \centering
   \includegraphics[width=0.95\linewidth]{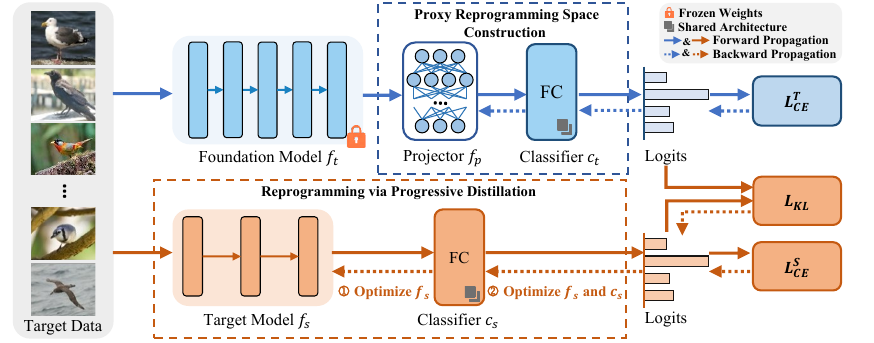}
   \caption{An overview of the proposed \textbf{T}ask-\textbf{D}riven \textbf{M}odel \textbf{R}eprogramng (\model). 
   We first reprogram a proxy space with the foundation model. The proxy space employ the projector and task-aligned classifier to alleviate the mismatch in tasks and the domain inconsistency.
   Then we reprogram the target model through the proxy space via progressive distillation in the feature extractor level and in the global level, which makes the student efficiently learn the knowledge extracted from the teacher.}
   \label{fig:method}
\end{figure*}

\section{Reprogramming with Foundation Models}
\subsection{Problem Statement}
\label{mthd:definition}
Let $f_t$ denote the vision foundation model (the teacher), which often has high capacity due to its tremendous parameters.
Note that, the broad data is usually unavailable in practice.
Given a set of target downstream data $\mD_{tgt}\small{=}\{\bx_i^{s},y_{i}\}_{i=1}^{N}$, directly deploying the foundation model $f_t$ to handle $\mD_{tgt}$ is computationally prohibitive and thus may be unacceptable in real-world applications.
To address these issues, we can employ a lightweight target model $f_s$ (the student) instead of the foundation model $f_t$ for deployment, and empower the relevant ability of $f_t$ to $f_s$.
However, it is non-trivial to achieve this goal due to the differences between the architectures of foundation and target models, the domains of pretraining and downstream data and the prototypes of pretraining and downstream tasks.

Recently, plenty of transfer learning methods~\cite{pan2009survey, chen2022model} have been proposed to reuse the same model from the original domain or task to new ones.
Nonetheless, these methods deviate from the intrinsic conditions in this scenario.
In addition, ordinary knowledge distillation methods assume that the training data and task prototypes of the teacher and student are the same~\cite{hinton2015distilling,tian2019contrastive,yang2022masked,touvron2021training,ji2021show,chen2021reviewkd}.
When directly applying ordinary knowledge distillation methods to the scenario with the domain differences, we find severe performance degradation, (see results in Tables~\ref{tab:swin},~\ref{tab:r50} and \ref{tab:256}, and analysis in Section~\ref{exp:sota}).

In this paper, we propose a new task-driven model reprogramming (\model) paradigm, to transfer a foundation model to a given downstream task that may have inconsistent domains, tasks and architectures with the foundation model.  
As shown in Figure~\ref{fig:method}, \model includes two key modules.
1) \textbf{Proxy reprogramming space} (c.f. Section~\ref{mthd:stage1}): we construct a proxy reprogramming space with the foundation model to project its knowledge/features.
2) \textbf{Progressive distillation} (c.f. Section~\ref{mthd:stage2}): we reprogram the target model via progressive distillation with the knowledge from the proxy space.
The overall training method of our \model is summarized in Algorithm~\ref{alg:method}.

\begin{algorithm}[h]
\caption{Task-Driven Model Reprogramming.}
\label{alg:method}
    \begin{algorithmic}[1]
    	\REQUIRE
    	Target data $\mD_{tgt}\small{=}\{\bx_i^{s},y_{i}\}_{i=1}^{N}$, a pretrained vision foundation model extractor $f_t$.  \\
    	\STATE \emph{// Stage 1: Proxy reprogramming space construction
     }
    	\STATE Randomly initialize a target extractor $f_s$ and classifier $c_s$ and a teacher classifier $c_t$. \emph{// $c_s$ and $c_t$ have the same architecture}
    	\STATE Randomly initialize a projector $f_p$ between $f_t$ and $c_t$.
        \FOR{a batch $\mX$ in $\mD_{tgt}$}
        \STATE Optimize $\min_{{f_p, c_t}} \mL^T_{ce}$ in Eqn.~(\ref{eq:ce_t}) on the batch $\mX$.
        \ENDFOR
    	\STATE \emph{// Stage 2: Reprogram via Progressive Distillation}
    	\STATE Set the student classifier $c_s\small~{:=}~c_t$. 
        \FOR{a batch $\mX$ in $\mD_{tgt}$}
        \STATE Optimize $\min_{{f_s}} \mL^S_{kd}$ in Eqn.~(\ref{eq:kd_1}) on the batch $\mX$.
        \ENDFOR
        \FOR{a batch $\mX$ in $\mD_{tgt}$}
        \STATE Optimize $\min_{{f_s,c_s}} \mL^S_{kd}$ in Eqn.~(\ref{eq:kd_1}) on the batch $\mX$.
        \ENDFOR
    	\STATE \textbf{Return} the target model extractor $f_s$ and classifier $c_s$.
    \end{algorithmic}
\end{algorithm}

\subsection{Proxy Reprogramming Space Construction}
\label{mthd:stage1}
As the foundation model is pretrained on broad data collected from wild domains under the conservative pretext tasks, how to break through the task mismatch and domain differences between the vision foundation model and the downstream target model is the key challenge. Here, we propose to construct a proxy reprogramming space with the foundation model to reconcile this potential conflict.
We comprehensively consider the teacher model, the student model and the target data domain when we design our proxy reprogramming space.
Specifically, in this stage, we maintain the foundation model $f_t$ as a powerful feature extractor frozen.
The frozen $f_t$ maintains a powerful and general feature extraction capability without bringing the computational cost of learning.
Then, we introduce the proxy reprogramming space to project the knowledge from the foundation model targeted to the downstream task. 
Concretely, we use a classifier as used in the downstream target model to align the task prototype. 
This classifier is directly related to the downstream task.
It is very suitable for target data, and it contains some data domain information, such as extracted feature dimension and categorical space.
By conducting the same classification task with the downstream data in the proxy space, we can mostly reprogram the foundation model to extract the downstream-aware knowledge, which helps alleviate the adverse effect induced by the task mismatch and domain difference.

There is still a gap
between the teacher extractor and the classifier.
When we use the target data $\bx_s$ as inputs, the fixed feature extractor $f_t$ is not fully applicable since the domain gap makes extracted features inaccurate and ineffective.
The $f_t$ is for the source domain, used to draw on the knowledge of the source domain and its own capabilities.
The $c_t$ is for the target domain, used to make the reprogrammed model better suited for downstream tasks.
We want to maintain both their strengths while aligning the domain, 
so we introduce a projector $f_p$ to transform the feature from $f_t$ into a refined counterpart.
The projector further converts the features extracted by the $f_t$ into the target domains and sends them into the $c_t$. Under the impetus of the front and rear ends, the projector is prompted to break through the data domain differences.
In this way, we use downstream tasks to drive model reprogramming in preparation for subsequent knowledge distillation.
On the principle of simplicity and adaptability, we use a backbone block of the teacher as the projector.

Formally, given the downstream data $\mD_{tgt}$, the formulation of reprogramming can be expressed as follows, 
\begin{align}
    \label{eq:ce_t}
    \mL^T_{ce}=\mmE_{(\bx^s, y)\sim \mD_{tgt}}\left[\CE(c_t(f_p(f_t(\bx^s))), y)\right],
\end{align}
where $f_p(\cdot)$ and $c_t(\cdot)$ are the neural module and the classifier respectively, which we should learn in this stage and are usually with a few parameters.  $\CE(\cdot)$ denotes the cross-entropy loss function to guide the reprogramming process.

\textbf{Generalization Analysis of Reprogramming.}
In this part, we discuss the necessity to reduce the domain inconsistency between the foundation model and the target model. As the study in\cite{ben2006analysis, long2015learning, feng2021kd3a}, the generalization bound for the domain adaptation between two domains dominantly depends on some distance between these two domains, \eg $\mH$-divergence \cite{ben2006analysis} and \textit{maximum mean discrepancy} (MMD) \cite{long2015learning, feng2021kd3a}.  
Recall $\mathbb{D}_{{S}}$ denotes the domain of downstream target data, let $\epsilon_{\mathbb{D}_{S}}(h)=\operatorname{Pr}_{(\mathbf{x}, \mathbf{y}) \sim \mathbb{D}_{S}}[h(\mathbf{x}) \neq \mathbf{y}]$ be the task risk with the model $h$,  $\mathbb{D}_{{T}}$ be the domain of broad data and assume the domain $\mathbb{D}_{S}$ shares the same task with the domain $\mathbb{D}_{T}$. We describe the generalization bound of the domain adaptation following \cite{ben2006analysis} as:
\begin{thm}
\label{thm1}
    Let  $\mathcal{H}$  be the model space,  $\epsilon_{\mathbb{D}_{{S}}}(h)$, $\epsilon_{\mathbb{D}_{T}}(h)$  be the task risks of the domain  $\mathbb{D}_{{S}}$  and  $\mathbb{D}_{T}$, respectively. Then  $\forall h \in \mathcal{H}$, we have:
    \begin{equation}
        \epsilon_{\mathbb{D}_{S}}(h) \leq \epsilon_{\mathbb{D}_{{T}}}(h)+d_{\mathcal{H}}\left(\mathbb{D}_{{S}}, \mathbb{D}_{T}\right)+C_{0},
    \end{equation}
    where $C_0$ is a constant for the complexity of the model space and $d_{\mathcal{H}}$ is the $\mH$-divergence between $\mathbb{D}_{{S}}$ and $\mathbb{D}_{T}$, which is defined as:
    \begin{equation}
    \label{d_H}
    \begin{aligned}
        d_{\mathcal{H}}(\mathbb{D}_{{S}}, \mathbb{D}_{T}) 
        &= 2 \sup _{\eta \in \mathcal{H}} \Big|\operatorname{Pr}_{\mathbf{x}^{s} \sim \mathbb{D}_{{S}}}\left[\eta\left(\mathbf{x}^{s}\right)=1\right] \\
        &- \operatorname{Pr}_{\mathbf{x}^{t} \sim \mathbb{D}_{T}}\left[\eta\left(\mathbf{x}^{t}\right)=1\right]\Big|.        
    \end{aligned}
    \end{equation}
    where $\eta \in \mH$ can be viewed as a two-sample classifier, \eg a (kernel) Parzen window classifier \cite{mansour2009domain} and a deep-kernel MMD-based two-sample classifier \cite{liu2020learning}. 
\end{thm}
Theorem \ref{thm1} suggests a lower domain discrepancy between the two domains $d_{\mathcal{H}}(\mathbb{D}_{{S}}, \mathbb{D}_{T})$ would have the tight generalization bound. This can be achieved by learning the proper representations of samples from these two domains according to Eqn. (\ref{d_H}). In our method, we introduce a domain adapter via reprogramming to learn a proxy space of two domains for better alignment using Eqn. (\ref{eq:ce_t}). We provide the empirical study in Section \ref{exp:domaingap} to validate this. Moreover, if some broad data is available, we can use MMD between $\mathbb{D}_{{S}}$ and $\mathbb{D}_{T}$ as auxiliary supervision in Eqn. (\ref{eq:ce_t}), which further reduces the domain gaps and makes the target models achieve better performance (see Table~\ref{tab:mmd}).

\subsection{Reprogramming via Progressive Distillation}
\label{mthd:stage2}
With the ``emergence and homogenization"~\cite{bommasani2021opportunities} in the previous stage, the goal of downstream task-driven model reprogramming in this stage is to distill the parts relevant to downstream tasks from the hidden huge knowledge of the foundation model to the target model. 
Distilling the parts relevant to downstream tasks from the hidden huge knowledge of the foundation model and passing them to the target model is a key step to achieve the goal.
The proposed downstream task-driven reprogramming already provides a good basis for distillation, since we have sent the classifier to the teacher model ahead of time to learn.
After completing the training of the teacher model, the classifier $c_t$ has been fully optimized under the powerful feature extractor and the scene data.
Here, we propose to reprogram the target model via progressive distillation from the proxy space.
We first reprogram the target classifier by transferring the task-aligned classifier from the proxy space. Note that, the task-aligned classifier $c_t$ has the same structure as the student classifier $c_s$ and thus we directly copy $c_t$ to $c_s$. This provides a good starting point in the approximately same space to efficiently conduct the knowledge transfer.
The task-aligned classifier in the proxy space captures both the knowledge prior of the foundation model and the knowledge of the target data, and the transfer process reprograms the outcomes to the target model.

The progressive distillation consists of \emph{extractor knowledge distillation} and \emph{global knowledge distillation}.
In the extractor knowledge distillation, we fix the student classifier $c_s$ and only perform knowledge distillation training on the student feature extractor $f_s$.
Through this, $f_s$ learns to approach the foundation model and minimize the gap induced by the model structure. Formally, it is expressed by the following equation,
\begin{equation} \label{eq:kd_1}
\begin{aligned}
    \mL^S_{kd}=\mmE_{(\bx^s, y)\sim\mD_{tgt}} [&~\KL(c_s(f_s(\bx^s)), M(\bx^s)) \\
    + &~\CE(c_s(f_s(\bx^s)), y)],
\end{aligned}
\end{equation}
where $\KL(\cdot, \cdot)$ is Kullback-Leibler divergence and $M$ is the function composition of the pretrained and frozen components $f_t$, $f_p$ and $c_t$.
After this, we perform the \emph{global knowledge distillation} as the second step, which minimizes the $\mL^S_{kd}$ in Eqn.~(\ref{eq:kd_1}) to finetune $f_s$ and $c_s$. This step refines the compatibility of two modules and help to improve the performance of the training. In summary, the proposed progressive cross-domain distillation through proxy space transfer makes full use of the advantages established in the reprogramming stage and helps the student efficiently learn the knowledge extracted from the teacher model.

\subsection{Wide Applicability of Reprogramming}
Our proposed \model is applicable to various scenarios with \textit{unavailable broad data}, and \textit{limited target data}, which is very practical for the applications of foundation models.

\textbf{Applicability to different pre-trained model types}.
It is worth noting that \model is independent of the architecture of the teacher model.
Specially, the recent vision foundation models mostly adopt Transformer~\cite{dosovitskiy2020image,liu2021swin,peng2023conformer} architectures due to their potential of pretraining on broad data.
This makes the early feature-based knowledge distillation methods~\cite{tian2019contrastive,yang2022masked} hard to handle differences between the architectures. 
Our proposed method is able to avoid this by distillation from the proxy reprogramming space instead of the feature space, so it is compatible with CNN, transformer or their mix.

\textbf{Applicability to unavailable broad data}.
As mentioned in Section~\ref{mthd:definition}, the massive broad data of the foundation model is often unavailable due to its high commercial values.
Besides, the foundation model often requires significant computational resources while pretraining.
In this sense, such models would also be not released publicly.
Foundation models (\eg GPT-3) usually only provide external services in the form of interfaces.
In this case, we can only obtain the outputs of the whole foundation model.
Even without access to the broad data and the foundation model, \model is able to learn a promising target model via the proposed reprogramming scheme.

\textbf{Applicability to limited target data}.
In practice, we may only obtain limited target data due to the high annotation cost.
In this case, the target model would only achieve poor performance by training from scratch.
Fortunately, we have a well-developed foundation model pretrained on massive data in the wild domain.
Note that the foundation model learns rich knowledge from the broad data and may well generalize to various downstream tasks.
In this sense, it is possible for the foundation model learned on the limited target data to annotate high-quality pseudo labels for unlabeled data.
With both the limited labeled data and the pseudo-label data, our proposed \model is still able to learn a target model with competitive or even better performance (see results and analysis in Table~\ref{tab:semi}).

\section{Experiments}\label{exp}
We conduct extensive experiments on a range of datasets and model architectures. Then, we validate our framework under semi-supervised learning.
Next, more comprehensive experiments are conducted to understand the intrinsic factors and limitations. Furthermore, we analyse the effectiveness of our framework by ablation experiments.
The source code will be released after the paper is published.

\begin{table*}[]
\centering
\caption{Comparisons with the state-of-the-art knowledge distillation methods in top-1 Accuracy (\%). In the experiments, we adopt Swin-B as the teacher model, which is pretrained on ImageNet-1k and achieves 83.1\% accuracy. We highlight the highest and the second highest results in bold and underlined, respectively. ``Baseline'' denotes that the student models are trained without knowledge distillation.}
\setlength{\tabcolsep}{1.45mm}
\renewcommand\arraystretch{1.3}
{
\begin{tabular}{c|cc|ccccc|p{1.5cm}<{\centering} >{\columncolor{pink!25}}c}
\hline
     Dataset      & Student & Baseline & {\makecell[c]{CRD~\cite{tian2019contrastive}}}& {\makecell[c]{AFD~\cite{ji2021show}}} & {\makecell[c]{ReviewKD~\cite{chen2021reviewkd}}} & {\makecell[c]{MGD~\cite{yang2022masked}}}&{\makecell[c]{REFILLED~\cite{ye2020distilling}}}& Tea.  & \model(Ours) \\ \hline
\multirow{3}{*}{CUB-200} &   ResNet-18  & 59.47 &  \underline{62.04} & 32.38 & 58.99 & 40.43 & 54.36 & 81.73 &\textbf{71.98} \\ 
                    &  MobileNetV2    & 62.90 & \underline{65.48} & 35.76 & 49.03 & 54.29 & 50.55 & 80.92 & \textbf{76.04} \\
                  &   ShuffleNetV2    & 58.88 & \underline{60.54} & 25.23  & 41.81 & 46.54 & 42.92 & 81.32 & \textbf{71.09} \\
                  \hline
\multirow{3}{*}{Oxford-102} &  ResNet-18 & 47.12 & \underline{53.16} & 36.62 & 41.32 & 37.84&  51.56 & 89.57 & \textbf{67.24} \\ 
                            &  MobileNetV2 & 37.94 & 56.96 & 37.42  & 43.55 & 45.64 & \underline{57.11} & 88.06 & \textbf{73.65} \\
                            &  ShuffleNetV2 & 50.54 & \underline{51.21} & 32.92& 37.71 & 36.51 & 44.18 & 88.68 & \textbf{71.32} \\
                  \hline
\multirow{3}{*}{Stanford Dogs} & ResNet-18 & 56.46  & 62.29 & 52.09 & \underline{66.98} & 55.36 & 65.57& 94.45 & \textbf{67.57}\\
                                & MobileNetV2 & 61.57 & 65.15 & 57.14 & 60.18 & 59.21 & \underline{67.86} & 94.53 & \textbf{67.91} \\
                                & ShuffleNetV2 & 56.04  & \underline{59.17} & 44.25& 54.32 & 50.54  & 58.28& 94.48 & \textbf{64.02} \\
                  \hline
\multirow{3}{*}{Oxford-IIIT-Pet} & ResNet-18 & 69.45 & \underline{75.54} & 61.39 & 68.55 & 55.67 & 74.59 & 93.23 & \textbf{78.47}\\
                                & MobileNetV2 & 63.50  & 75.00 & 63.64 & 63.79 & 63.66 & \underline{76.65} & 93.15 & \textbf{80.97} \\
                                & ShuffleNetV2 &  67.63  & 70.78 & 50.92 & 54.45 & 56.25  & \underline{72.28} & 93.26 & \textbf{76.44} \\
\hline
\end{tabular}
}
\label{tab:swin}
\end{table*}
 
\begin{table*}[]
\centering
\caption{Comparisons with the state-of-the-art knowledge distillation methods in top-1 Accuracy (\%). In the experiments, we adopt ResNet-50 as the teacher model, which is pretrained on ImageNet-1k and achieves 77.76\% accuracy.}
\setlength{\tabcolsep}{1.45mm}
\renewcommand\arraystretch{1.3}
{
\begin{tabular}{c|cc|ccccc|p{1.5cm}<{\centering} >{\columncolor{pink!25}}c}
\hline
     Dataset      &Student & Baseline & {\makecell[c]{CRD~\cite{tian2019contrastive}}}& {\makecell[c]{AFD~\cite{ji2021show}}}  & {\makecell[c]{ReviewKD~\cite{chen2021reviewkd}}} & {\makecell[c]{MGD~\cite{yang2022masked}}}& {\makecell[c]{REFILLED~\cite{ye2020distilling}}}  & Tea. & \model(Ours) \\ \hline

\multirow{3}{*}{CUB-200}  &   ResNet-18 & 59.47  & \underline{64.18} & 48.52 & 60.63 & 46.47& 54.22 &  71.00& \textbf{71.48} \\ 
                        & MobileNetV2    & 62.90  & \underline{65.67} & 41.00 & 51.72 & 58.33 & 49.11& 71.81 & \textbf{72.60} \\
                        &   ShuffleNetV2   & 58.88 & \underline{61.99}& 39.02  & 46.06 & 54.17& 45.46 & 71.67 & \textbf{68.69} \\
                
                \hline
\multirow{3}{*}{Oxford-102} &  ResNet-18 & 47.12 & \underline{51.81} & 39.11 & 43.93 & 39.33& 51.78  & 76.98 & \textbf{55.40} \\
                            & MobileNetV2   & 37.94  & 48.31 & 46.19 & 46.08 &46.67 & \underline{57.91}& 77.05 & \textbf{61.57} \\
                            &  ShuffleNetV2   & 50.54 & 47.71& 37.57  & 35.69 & 40.60 & \underline{50.82} & 76.63 & \textbf{61.70} \\
             \hline
\multirow{3}{*} {Stanford Dogs}& ResNet-18 & 56.46 & 65.18& 59.92  & 66.81 & 62.17 & \underline{66.92} & 88.32 & \textbf{68.62} \\
                                & MobileNetV2 & 61.57 & \underline{68.01} & 66.93  & 58.84 & 65.39 & 63.01 & 88.43 & \textbf{71.31} \\
                                & ShuffleNetV2  & 56.04 & 60.82& 54.90  & 51.01 & 57.56 & \underline{63.14} &  88.56 &\textbf{64.03} \\
               \hline
\multirow{3}{*} {Oxford-IIIT-Pet}& ResNet-18 & 69.45 & \underline{75.78} &  72.04  & 72.78 &  61.41 & 74.04 & 92.33 & \textbf{80.51} \\
                                & MobileNetV2 & 63.50 & \underline{75.43} &  70.35  & 62.88 & 69.10 &  73.69 & 92.60 & \textbf{82.82} \\
                                & ShuffleNetV2  & 67.63 & 71.82 &  66.77  &  56.14 &  63.26 & \underline{72.36}  & 92.50 & \textbf{78.64} \\
\hline
\end{tabular}
}
\label{tab:r50}
\end{table*}

\begin{table*}[]
\centering
\caption{Comparisons with the state-of-the-art knowledge distillation methods in terms of top-1 Accuracy (\%) on Caltech-256-60. We report the performances of students when adopting ResNet-50 and Swin-B, respectively.}

\setlength{\tabcolsep}{1.8mm}
\renewcommand\arraystretch{1.3}
{
\begin{tabular}{c|cc|ccccc|p{1.5cm}<{\centering} >{\columncolor{pink!25}}c}
\hline
     Teacher      & Student & Baseline & {\makecell[c]{CRD~\cite{tian2019contrastive}}}& {\makecell[c]{AFD~\cite{ji2021show}}} & {\makecell[c]{ReviewKD~\cite{chen2021reviewkd}}} & {\makecell[c]{MGD~\cite{yang2022masked}}} &{\makecell[c]{REFILLED~\cite{ye2020distilling}}} & Tea. & \model(Ours) \\ \hline
\multirow{3}{*}{Swin-B} &  ResNet-18 & 49.18 & \underline{53.81} & 40.39 & 52.98 & 37.64 & 53.67 & 89.24 & \textbf{57.45} \\ 
                        &  MobileNetV2   & 50.68 & \underline{56.92} & 44.49 & 46.98 & 46.07 & 55.31 & 83.28 & \textbf{59.49} \\
                        &  ShuffleNetV2 & 47.68 & \underline{50.31} & 31.56 & 41.75 & 42.37 & 48.38 & 83.22 & \textbf{53.32} \\
\hline
\multirow{3}{*}{ResNet50} &   ResNet-18  & 49.18 & \underline{55.31} & 48.64 & 53.98 & 43.03 & 53.93 & 82.21 & \textbf{58.46} \\ 
                            &  MobileNetV2    & 50.68 & \underline{59.70} & 47.90 & 50.60 & 50.19 & 55.40 & 83.28 & \textbf{62.66} \\
                            &   ShuffleNetV2    & 47.68 & \underline{51.71} &  36.50 & 41.10 & 46.94 & 51.42 & 83.22 & \textbf{54.78} \\
                  \hline

\end{tabular}
}
\label{tab:256}
\end{table*}

\subsection{Experimental Setup}
\noindent\textbf{Datasets.} 
We use ImageNet~\cite{deng2009imagenet} to train the vision foundation models, and consider four downstream tasks from different domains: CUB-200~\cite{wah2011caltech}, Oxford-102~\cite{nilsback2008automated} Stanford Dogs~\cite{KhoslaYaoJayadevaprakashFeiFei_FGVC2011} and Oxford-IIIT-Pet~\cite{parkhi12a}. We also verify \model on Caltech-256-60~\cite{griffin2007caltech} as a task from wilder domains.

\noindent\textbf{Baselines.} The state-of-the-art distillation methods including CRD~\cite{tian2019contrastive}, ReviewKD~\cite{chen2021reviewkd}, MGD~\cite{yang2022masked}, AFD~\cite{ji2021show}, and a cross-task distillation method REFILLED~\cite{ye2020distilling} are used as the baselines to compare with our \model. Note that, as the logit-based distillation methods cannot distill a foundation model into a domain-different (e.g., category-different) downstream target model, we do not compare them with our TDMR. For a comprehensive comparison, we also compare \model with MRKD which means a simple combination of model reprogramming (MR) and knowledge distillation (KD), and transfer learning method linear probing~\cite{he2022masked,radford2021learning}.

\noindent\textbf{Implementation details.} Following\cite{tian2019contrastive}, we adopt the SGD optimizer with the momentum 0.9 and the weight decay $5\times10^{-4}$ during training. 
We employ CosineAnnealingLR as the learning rate scheduler for both stages. For the first stage of \model, we set the initial learning rate to 0.01 and train the projector $f_p$ and the classifier $c_t$ for 150 epochs. In the second stage, we train students for 400 epochs. For the teacher Swin-B, we set the learning rate of the student $f_s$ to 0.1. For the teacher ResNet-50, we set the learning rate of the student $f_s$ to 0.05. For the comparison methods, we obtain the results by means of their official publicly released code with setting the training epochs to 400 for a fair comparison.
Unless specifically stated, the architecture of $f_p(\cdot)$ is one block structure of the teacher model.
We align all training settings in implementing the baselines. 

\noindent\textbf{The Accuracy of Teacher Models.} We provide the top-1 accuracy of the teacher model after applying downstream task-driven reprogramming. For simplicity, we abbreviate the reprogrammed teacher as ``Tea.'' in Table~\ref{tab:swin},~\ref{tab:r50},~\ref{tab:256},~\ref{response:tab:adaptformer},~\ref{supp:tab:vfm} and~\ref{supp:tab:proj} for readers' reference. 
Note that the ``Tea.'' for each student is different as we reprogram the teacher model according to the specific downstream tasks and 
architectures of the student model.
In the experiments, Swin-B and ResNet-50 are pretrained on ImageNet-1k and respectively achieves 83.1\% and 77.76\% in accuracy.

\subsection{Comparisons on Downstream Tasks}
\label{exp:sota}
\noindent\textbf{Comparisons with state-of-the-art distillation methods.}
In this section, we consider a popular vision foundation model, Swin transformer (Swin-B)~\cite{liu2021swin} pretrained on ImageNet-1k as the teacher model, and employ ResNet-18~\cite{he2016deep}, MobileNetV2~\cite{sandler2018mobilenetv2} and ShuffleNetV2~\cite{Ma_2018_ECCV} as students. The results are summarized in Table~\ref{tab:swin}. 
As can be seen, our method outperforms other distillation methods by a large margin. For example, \model achieves 76.04\% accuracy for MobileNetV2 on CUB-200, which obtains 10.56\% performance improvement over CRD~\cite{tian2019contrastive} (65.48\%) and 13.14\% performance improvement over the basic student model (62.90\%) in terms of top-1 accuracy. 
Note that, the dataset used to validate our method is aligned with a certain downstream task, and
the domain in label space is different from the pre-training dataset. The experimental results confirm the wide applicability of our method. Especially, our method is capable of working effectively even when there are significant architectural disparities between the teacher model (based on transformer) and the student model (based on CNN).

Considering that convolutional architectures are also commonly applied in some vision foundation models, we also conduct experiments using ResNet-50~\cite{he2016deep} pretrained on ImageNet-1k as the teacher model.
As shown in Table~\ref{tab:r50}, the same conclusion that our method performs better than other baselines is also held. Especially, our method exhibits superior performance compared to CRD~\cite{tian2019contrastive} on the ShuffleNetV2 model for the Oxford-102 dataset, achieving a performance gain of 13.86\%.
To further verify the effectiveness of our method, we conduct experiments on another wilder domain dataset  Caltech-256-60~\cite{griffin2007caltech} in Table~\ref{tab:256}.
As can be seen, \model similarly achieves superiority when applying different teacher and student models. Specifically, \model outperforms CRD~\cite{tian2019contrastive} by 3.64\% and MGD~\cite{yang2022masked} by 19.81\% in top-1 accuracy over ResNet-18 when employing Swin-B as the teacher.

\begin{table}[t]
  \centering
  \caption{Comparison with ``model reprogramming + knowledge distillation'' (MRKD) that uses adaptformer~\cite{chen2022adaptformer} with vanilla KD. ``Tea.'' is the reprogrammed Swin-B.}
  \setlength{\tabcolsep}{2.6mm}
  \renewcommand\arraystretch{1.3}
  {
    \begin{tabular}{c|c|cc|cc}
    \hline
    \multirow{2}{*}{Dataset}     & \multirow{2}{*}{Student} & \multicolumn{2}{c|}{MRKD}                       & \multicolumn{2}{c}{TDMR}           \\ \cline{3-6} 
                                 &                          & \multicolumn{1}{c|}{Tea.}                   & Stu.  & \multicolumn{1}{c|}{Tea.}  & Stu.  \\ \hline
    \multirow{3}{*}{Caltech-256} & ResNet-18                & \multicolumn{1}{c|}{\multirow{3}{*}{86.34}} & 50.11 & \multicolumn{1}{c|}{89.24} & 57.45 \\ \cline{2-2} \cline{4-6} 
                                 & MobileNetV2              & \multicolumn{1}{c|}{}                       & 54.12 & \multicolumn{1}{c|}{83.28} & 59.49 \\ \cline{2-2} \cline{4-6} 
                                 & ShuffleNetV2             & \multicolumn{1}{c|}{}                       & 49.31 & \multicolumn{1}{c|}{83.22} & 53.32 \\ \hline
    \multirow{3}{*}{Oxford-102}  & ResNet-18                & \multicolumn{1}{c|}{\multirow{3}{*}{88.29}} & 52.38 & \multicolumn{1}{c|}{89.57} & 67.24 \\ \cline{2-2} \cline{4-6} 
                                 & MobileNetV2              & \multicolumn{1}{c|}{}                       & 59.42 & \multicolumn{1}{c|}{88.06} & 73.65 \\ \cline{2-2} \cline{4-6} 
                                 & ShuffleNetV2             & \multicolumn{1}{c|}{}                       & 57.58 & \multicolumn{1}{c|}{88.68} & 71.32 \\ \hline
    \end{tabular}
    }
    \label{response:tab:adaptformer}
\end{table}

\begin{figure*}[t]
    \begin{minipage}[t]{0.32\linewidth}
        \centering
        \includegraphics[width=1.0\linewidth]{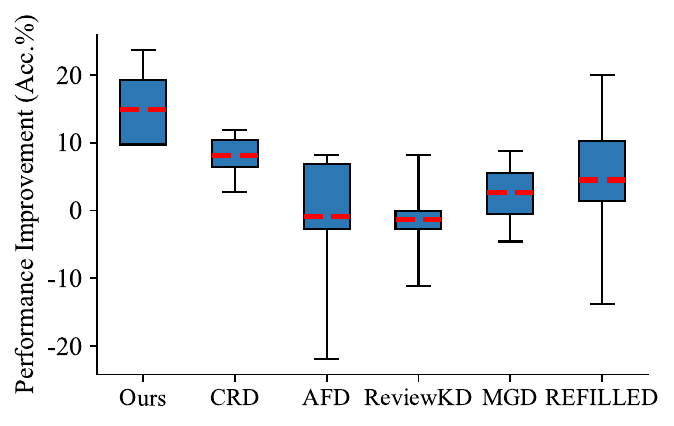}
        \caption{
        An illustration of different methods regarding the accuracy improvement across 5 considered datasets with box plots. 
        }
        \label{fig:accimprov}
    \end{minipage}%
    \hfill
    \begin{minipage}[t]{0.32\linewidth}
        \centering
        \includegraphics[width=1.0\linewidth]{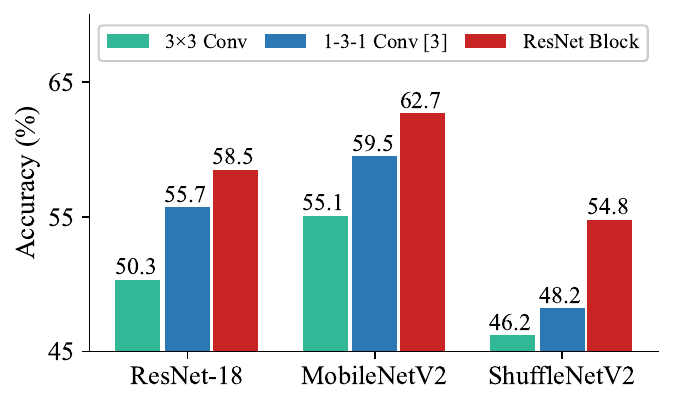}
        \caption{Comparisons of different projector architectures on Caltech-256-60.}
        \label{fig:projector}
    \end{minipage}
    \hfill
    \begin{minipage}[t]{0.32\linewidth}
        \centering
        \includegraphics[width=1.0\linewidth]{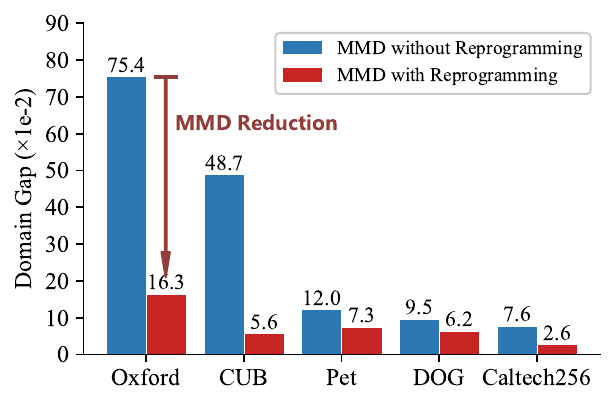}
        \caption{
        An illustration of domain gap (MMD) reduction between source data and target data \textbf{with/without} reprogramming.
        }
        \label{fig:mmd}
    \end{minipage}  
\end{figure*}

\noindent\textbf{On the stability of different distillation methods.}
From the above results, we find some distillation methods may improve student performance slightly, even degenerate the performance in some cases. 
To study the stability of each method, we compare the means and extremums of accuracy improvement over \textit{baseline} of each method on all datasets and draw the box-plot of the results in Figure~\ref{fig:accimprov}. 
From the figure, we see that \model has the highest mean improvement with relatively small fluctuations. Meanwhile, other methods, \eg AFD~\cite{ji2021show}, cannot guarantee consistent improvements for different domains, as it is infeasible to directly align features of the teacher and student model from different domains. Note that REFILLED~\cite{ye2020distilling} also does not show significant performance. This is probably because it cannot deal with the quite different relationships between samples from two domains very well. 
Unlike these methods, our method (\model) comprehensively considers the gaps in data and model architecture and efficiently transfers knowledge from the foundation model to the target model.

\noindent\textbf{Comparisons with ``MRKD''.}
To further demonstrate the effectiveness of \model, we compare \model with a direct combination of model reprogramming (MR) and knowledge distillation (KD) denoted by ``MRKD''. MRKD first utilizes a typical model reprogramming method adaptformer~\cite{chen2022adaptformer} to reprogram the teacher model. Then, it conducts vanilla knowledge distillation~\cite{hinton2015distilling} on the reprogrammed teacher.
We conduct experiments on the downstream Caltech-256 and Oxford-102 datasets with the same foundation model, \ie Swin-B pretrained on ImageNet-1k. Based on the results in Table~\ref{response:tab:adaptformer}, we can see TDMR significantly outperforms MRKD under different student architectures. For example, \model outperforms MRKD by 6.34\% in top-1 accuracy of ResNet-18 on Caltech-256. Comparing the results of ResNet-18 on Oxford-102, \model also achieves a 14\% better result than MRKD. These results provide strong support for the above advantages of our method.

We summarize three advantages of TDMR over MRKD: (i) In reprogramming phase, the model reprogramming in MRKD is to pursue the best reprogramming performance and does not consider KD, while we argue that the best teacher model might not be the best choice for the student model in KD.
Our model reprogramming in TDMR is to design a proper reprogramming space for KD, that is, a projector with a classifier is simple yet well-suited for the KD on downstream task.
(ii) In distillation phase, the KD in MRKD usually leverages one-stage distillation, a brute-forced transfer, while our KD in TDMR is a two-stage progressive distillation, a mild and more efficient way, which will be demonstrated in the following. (iii) More importantly, we reprogram students according to the proxy space, \ie we consider the knowledge of the reprogrammed teacher classifier, which is missing in MRKD.

\begin{table}[t]
  \centering
  \caption{Comparison with linear probing. We employ the pre-trained  ResNet-50 and Swin-B models as teachers and reprogram them by linear probing and our TDMR.}
  \setlength{\tabcolsep}{2.6mm}
  \renewcommand\arraystretch{1.3}
  {
  \begin{tabular}{c| c|c c| c c}
    \hline
    \multirow{2}{*}{Dataset} & \multirow{2}{*}{Student} & \multicolumn{2}{c|}{Swin-B} & \multicolumn{2}{c}{ResNet-50} \\ 
    & & Lin. & Ours& Lin.&Ours \\
    \hline
    \multirow{3}{*}{Caltech-256}&ResNet-18  &55.46 & \textbf{57.45}&58.34 &\textbf{58.46}\\
    & MobileNetV2& 59.08& \textbf{59.49} &62.23 & \textbf{62.66}\\ 
    &ShuffleNetV2  &53.17 & \textbf{53.32}&53.79 &\textbf{54.78}\\
     \hline
    \multirow{3}{*}{Oxford-102}&ResNet-18 &62.64 & \textbf{67.24} &55.30 &\textbf{55.84}\\ 
    & MobileNetV2 & 53.96& \textbf{73.65}&59.01 & \textbf{61.57}\\ 
    &ShuffleNetV2&65.52 & \textbf{71.32} &52.49 &\textbf{61.70} \\

    \hline
  \end{tabular}}
  
  \label{supp:tab:fc}
\end{table}

\noindent\textbf{Comparison with Linear Probing.}
For the foundation model application to downstream tasks, a common transfer method is the linear probing~\cite{radford2021learning,he2022masked} which just modifies the output dimension of the teacher classifier to the total number of categories of the target data. 
Then, the classifier is trained from scratch with the feature extractor frozen instead of fully fine-tuning the source model with high computation cost.
After conducting linear probing, the vanilla knowledge distillation~\cite{hinton2015distilling} method is used to train students with hard and soft labels.
The whole process above is denoted by ``Lin.''
We conduct experiments with various combinations of teachers and students on Caltech-256-60 and Oxford-102 to compare the common method with ours. From the results in Table~\ref{supp:tab:fc}, we can see that our method outperforms the common method ``Lin.''. 
Moreover, our method exhibits a better improvement on the Oxford-102 dataset than the Caltech-256 dataset. It can be attributed to the larger gap between the Oxford dataset and the pre-training dataset. 
These results provide further evidence of the superior performance of our proposed method, especially in scenarios with larger domain gaps.

\subsection{Effectiveness on Larger Models and Data}
\label{supp:sec:morefoundationmodels}
In this section, to further verify the effectiveness of our method on larger datasets and models, we employ more vision foundation models, \eg Swin-L~\cite{liu2021swin} and CLIP(ViT-B/32)~\cite{radford2021learning}, as teachers to. Swin-L is pretrained on ImageNet-21k. 
CLIP is pretrained on a large (image, text) pair dataset~\cite{radford2021learning}. 
We employ MobileNetV2 as the student and conduct experiments on Oxford-102. 
From the results in Table~\ref{supp:tab:vfm}, we find that our method is able to be applied to various vision foundation models with different numbers of parameters. Moreover, a student model can gain more improvement from a stronger teacher model.
To compare the transfer learning paradigm on larger data,
we also implement the ``Lin."~\cite{he2022masked} baseline to compare with TDMR. From the results, our method achieves a consistent gain under different settings, which shows the effectiveness and practical usefulness of TDMR.

\begin{table}[h]
\centering
\caption{Comparison with linear probing on larger models and data. We report the total number of parameters in different foundation models for a clear comparison.}
 \setlength{\tabcolsep}{0.9mm}  
  \renewcommand\arraystretch{1.3}
  {
\begin{tabular}{c|ccccc}
\hline
 Teacher     & ResNet-50 & Swin-B-1k &  Swin-B-21k  & Swin-L-21k &  CLIP  \\ \hline
\#Params (M) & 25    & 88     & 88    & 197    & 87 \\ 
 Tea.       & 77.05  & 88.06  & 99.57 & 99.51  & 92.55 \\
 Lin.       & 59.01  & 53.96  & 63.42 & 59.03  & 63.05 \\
TDMR        & \textbf{61.57}  & \textbf{73.65}  & \textbf{75.26} & \textbf{75.47}  & \textbf{76.77} \\ \hline
\end{tabular}
}
\label{supp:tab:vfm}
\end{table}

\subsection{Effectiveness under Semi-Supervised Learning}
\label{exp:semi-sup}
In this part, we show the effectiveness of our method extended for semi-supervised learning with some unlabeled data in the target data set. We conduct the experiment on Caltech-256. Specifically, we remove the labels of 90\% of training samples for each class.
Then, we perform the foundation model reprogramming with the remaining 10 percent of labeled samples.
In the progressive distillation stage, the supervision of the unlabeled samples in Eqn.~\eqref{eq:kd_1} is provided by vision foundation models. In total, we train the student model on the samples with ground-truth labels, pseudo labels and logits.
According to Table~\ref{tab:semi}, we can find that increasing the number of training samples with labels can increase the performance while increasing the number of unlabeled samples can also improve the performance comparably. Sometimes, it even performs better than the totally supervised counterpart (see the performance with Swin-B). This confirms that the foundation model reprogrammed by \model learns the domain-aware knowledge and provides reliable supervision to the downstream target model via our proposed progressive distillation method.
Note that, the experiments here are to validate the potential of TDMR in the semi-supervised scenarios by comparing the fully-supervised baseline, instead of proving that TDMR is better than other semi-supervised methods.

\begin{table}[]
\centering
  \caption{Comparisons under different numbers of labeled samples on Caltech-256. The numbers before and after the slash are the percentage of samples with ground-truth labels and pseudo labels per class compared to the number of original data, respectively.}
   \setlength{\tabcolsep}{1.8mm}
  \renewcommand\arraystretch{1.3}
  {
\begin{tabular}{c|c|ccc}
\hline
         Teacher                &     Student   &     10\% / 0\%        & 100\% / 0\% & 10\% / 90\% \\ \hline
\multirow{3}{*}{Swin-B}                &     ResNet-18        &             14.82         & 57.45 &  \textbf{59.74}\\
                                       &     MobileNetV2         &           18.75           & 59.49 & \textbf{61.82} \\
                                       &     ShuffleNetV2         &           13.28           & 53.32 & \textbf{56.03} \\ \hline
\multicolumn{1}{c|}{\multirow{3}{*}{ResNet-50}} & \multicolumn{1}{c|}{ResNet-18} & \multicolumn{1}{c}{15.44} & \textbf{58.46} & 55.79 \\
\multicolumn{1}{c|}{}                  & \multicolumn{1}{c|}{MobileNetV2} & \multicolumn{1}{c}{21.63} & \textbf{62.66} & 59.12 \\
\multicolumn{1}{c|}{}                  & \multicolumn{1}{c|}{ShuffleNetV2} & \multicolumn{1}{c}{17.04} & \textbf{54.78} & 54.64 \\ \hline
\end{tabular}
}
\label{tab:semi}
\end{table}

\subsection{Ablation Studies}
\label{exp:ablation}
In this section, we study the effect of different projector architectures and evaluate the effectiveness of the progressive knowledge distillation with the proxy space transfer.

\begin{table*}[h]
\centering
\caption{Comparison of different projectors. Here, we validate different projector structures on Swin-B and ResNet-50, and report the top-1 accuracy of the teacher and the student (in the ``Tea.'' column and the ``Student'' column), respectively.}
\resizebox{1\textwidth}{!}{
\tiny
\begin{tabular}{c|c|c|cc|cc|cc}
\hline
\multirow{2}{*}{Dataset} &\multirow{2}{*}{Teacher} & \multirow{2}{*}{Projector} & \multicolumn{2}{c|}{ResNet-18} & \multicolumn{2}{c|}{MobileNetV2} & \multicolumn{2}{c}{ShuffleNetV2} \\\cline{4-9}
    &              &                   &     Tea.      &     Student      &    Tea.       &      Student     &    Tea.       &      Student    \\ \hline
\multirow{6}{*}{Caltech-256-60}&\multirow{3}{*}{Swin-B} & MHSA &       89.04    &     56.82      &   88.52        &    59.18       &  88.63         &    52.62      \\
       &                                          & ViT Block  &     89.08      &     56.94      &   88.98        &    58.77       &  88.89         &    52.31      \\
       &                                          & Swin Block &     89.24      &     57.45      &   88.81        &     59.49      &  88.59         &   53.40       \\ \cline{2-9}
&\multirow{3}{*}{ResNet-50} & 3$\times$3 Conv            &   83.63        &     50.29      &    84.12       &     55.13      &    83.92       &     46.18     \\
        &          & 1-3-1 Conv~\cite{chen2022knowledge} &  84.74         &     55.71      &    84.53       &    59.51       &    85.11       &     48.15     \\
         &         & ResNet Block                        &  82.21         &     58.46      &    83.28       &     62.66      &    83.22       &     54.78     \\ \hline
\multirow{6}{*}{Oxford-102}&\multirow{3}{*}{Swin-B} & MHSA &   89.44        &    64.51       &   88.55        &    74.63       &   88.84        &   70.79       \\
       &                                      & ViT Block  &   89.16        &     66.72      &   88.38        &    75.63       &    88.74       &    70.95      \\
       &                                      & Swin Block &    89.57       &     67.24      &   88.06        &    73.65       &   88.68        &     71.32     \\ \cline{2-9}
&\multirow{3}{*}{ResNet-50} & 3$\times$3 Conv            &    79.60       &    52.25       &    79.86       &   53.82        &     79.89      &   47.48       \\
        &          & 1-3-1 Conv~\cite{chen2022knowledge} &    83.98       &    54.78       &    83.86       &   60.41        &     84.12      &   52.43       \\
         &         & ResNet Block                        &    76.98       &   55.40        &   77.05        &    61.57       &     76.63      &     61.70     \\ \hline
\end{tabular}}
\label{supp:tab:proj}
\end{table*}

\noindent
\textbf{Effect of the projector architecture.}
We study the effect of different architectures of the projector on Caltech-256-60 and Oxford-102. We plot the results of ResNet-50 in Figure~\ref{fig:projector} and the whole results can be seen in Table~\ref{supp:tab:proj}.
For ResNet-50 teacher model, we compare the ``ResNet Block" structure with ``3$\times$3 Conv" and ``1-3-1 Conv", where
``3$\times$3 Conv" denotes a standard sequence of 3$\times$3 convolutional layer, batchnorm layer and activation layer.  
``1-3-1 Conv" denotes 1$\times$1 Conv-3$\times$3 Conv-1$\times$1 Conv structure in~\cite{chen2022knowledge}. 
We can see that a single block structure of the teacher backbone achieves the best performance.
The reason may be that the projector that highly matches the model is its own structure that bridges the gaps effectively during the reprogramming process and help students reach better performance.
For Swin-B teacher model, we compare a ``Swin Block" structure with ``MHSA" and ``ViT Block".
``MHSA" denotes a standard multi-head self-attention structure.  
``ViT Block" denotes a single block structure of the Vision Transformer~\cite{dosovitskiy2020image}. The results are given in Table~\ref{supp:tab:proj}. We find that our method outperforms others in most cases and the same conclusion that taking an original block of the teacher model as the projector is good enough can be drawn.
This phenomenon brings great convenience to the applications of our method because we do not need to carefully design the projector structure. 
We simply use the off-the-shelf structures of teacher models as the projector and obtain superior performance.

\noindent
\textbf{Effect of the progressive distillation with the proxy space transfer.}
To verify the effectiveness of the progressive distillation strategy with proxy space transfer, we conduct a series of ablation experiments on Caltech-256-60. 
We employ ResNet-50 as the teacher and use three lightweight models as students. 
In Table~\ref{tab:cls-reuse}, we report the top-1 accuracy of student models with different distillation strategies.
We reprogram the teacher model in the same way (c.f. Section~\ref{mthd:stage1}) and set three kinds of distillation strategies.
1) Normal: we train the whole student model from scratch by minimizing the Eqn.\eqref{eq:kd_1} without proxy space transfer.
2) Proxy Trans.: we copy the parameters from the $c_t$ to the student classifier $c_s$ and only train the student extractor $f_s$ with the $c_s$ frozen.
3) Proxy Copy: we also copy the parameters of the teacher classifier $c_t$ to the student classifier $c_s$ but do not fix the student classifier in the whole training process.
4) Progressive: based on the proxy space transfer, we train the $f_s$ and $c_s$ progressively (c.f. Section~\ref{mthd:stage2}).

From the results reported in Table~\ref{tab:cls-reuse}, ``Progressive" achieves the best performance compared with the other settings, which suggests that progressive distillation with proxy space transfer enables more comprehensive training of student models for better performance.
``Proxy Trans.'' shows competitive performance compared with ``Normal'', which shows that the proxy space has learned effective knowledge during the reprogramming process, so it plays an important role even without training in stage 2. The reason that ``Proxy Trans.'' does not get obvious superior results is because that there may exist some mismatch between the $c_s$ and $f_s$. 
``Proxy Copy'' also obtains the relative improvement but is not better than ``Progressive". 
Therefore, we perform progressive distillation to not only retain the knowledge that $c_s$ has learned, but also further optimize the student at the global level.
The superior results of ``Progressive" show that the proxy space transfer provides a better basis for knowledge distillation.

\begin{table}[t]
  \centering
  \caption{Effect of different knowledge distillation strategies. We report top-1 accuracy (\%) on Caltech-256-60.}
  \setlength{\tabcolsep}{2.4mm}
  \renewcommand\arraystretch{1.3}
  {
  \small
  \begin{tabular}{c|c c c}
    \hline
    KD Strategy
    & ResNet18 
    & MobileNetV2
    & ShuffleNetV2
    \\ \hline
    Normal & 55.54 &61.69 & 53.38\\ 
    Proxy Trans. & 57.47 &60.85 & 52.80\\ 
    Proxy Copy   & 55.83 & 61.63 & 54.24 \\
    Progressive &\textbf{58.46}  &\textbf{62.66} &\textbf{54.78}  \\ \hline
  \end{tabular}
  }
  \label{tab:cls-reuse}
\end{table}

\subsection{Further Analysis of Domain Gap}
\label{exp:domaingap}
In this section, we evaluate the effectiveness of domain adapter without accessing broad data.
Then, we utilize the MMD between the broad data and the target data as additional supervision to narrow down the domain gap when the broad data is available.

\begin{table}[t]
  \centering
  \caption{Impact of the MMD loss. We use ResNet-50 and MobileNetV2 as the teacher and student respectively. We report top-1 accuracy (\%) and the value of MMD that is scaled up by 100.}
 \setlength{\tabcolsep}{1.9mm}
  \renewcommand\arraystretch{1.3}
  { \begin{tabular}{c|c|cc|cc}
    \hline
    \multirow{2}{*}{Teacher} & \multirow{2}{*}{\makecell[c]{MMD \\ Loss}} & \multicolumn{2}{c|}{Oxford-102} & \multicolumn{2}{c}{CUB-200} \\ \cline{3-6} 
                          &          &  Domain Gap   &   Acc.  &   Domain Gap     &  Acc.  \\ \hline
   \multirow{2}{*}{Swin-B}  &       \xmark    &   36.6   &   73.65   &  6.1   &  76.04 \\
                        &      \checkmark      &  \textbf{2.4}    &  \textbf{73.84}  &    \textbf{0.7}    &  \textbf{76.49}    \\ \hline
    \multirow{2}{*}{ResNet-50}  &     \xmark  &   16.3  &   61.57    &   5.6  &  72.60  \\
                      &      \checkmark       &   \textbf{6.0}   &   \textbf{61.97}  &   \textbf{3.5}   &    \textbf{73.16} \\ \hline
    \end{tabular}
  }
  \label{tab:mmd}
\end{table}

\noindent
\textbf{Impact of domain gap.}
To further understand the effectiveness of our domain adapter, we measure the domain gap between the broad data and the target data before and after constructing the proxy reprogramming space. Specifically, we use ResNet-50 to extract the last-layer feature of ImageNet validation set and the target data set (\eg Oxford-102, CUB-200, Oxford-IIIT-Pet, Stanford Dogs and Caltech-256-60), then calculate their MMD as the domain gap of two datasets. The results are reported in Figure~\ref{fig:mmd}.
The domain gap between the two datasets decreases when the teacher model is equipped with a domain adapter, which is consistent with the claim in Theorem~\ref{thm1}.

\noindent
\textbf{Combination with the MMD loss.}
If a part of the broad data is available, we can utilize the MMD loss between the broad data and the target data to explicitly regularize the training (c.f. Section~\ref{mthd:stage1}). To this end, we randomly select 10000 images in ImageNet training set to enlarge the training set for the domain adapter. In addition to Eq.~\eqref{eq:ce_t}, we select the same number of images from broad and target data to compute their MMD, which is added as an additional loss. 
In Table~\ref{tab:mmd}, we report the top-1 accuracy on two datasets and compare the domain gap with and without the MMD loss. From the results, the lower domain gap induced by the MMD loss indeed further improves the performance of knowledge transfer.

\section{Conclusion}\label{exp}
In this paper, we have proposed an efficient Task-Driven Model Reprogramming framework to realize the application of foundation models to various downstream tasks.
In \model, we constructed a proxy space with the foundation model to project the knowledge from it, which overcomes the adverse effect of task mismatch and domain inconsistency. Then, we construct a progressive distillation via the proxy space transfer to make the target model efficiently learn from the reprogrammed teacher model.
We have verified our methods on various downstream tasks and semi-supervised learning scene and the extensive results have demonstrated the effectiveness of \model. In the future, we can extend our framework for richer vision tasks like detection, segmentation and image captioning to validate its effectiveness in a more general scope.

\appendices

\bibliographystyle{IEEEtran}
\bibliography{TDMR}

%
\begin{IEEEbiography}[{\includegraphics[width=1in,height=1.25in,clip,keepaspectratio]{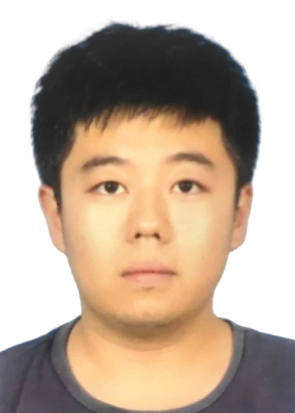}}]{Shoukai Xu}
is currently a Ph.D. candidate at South China University of Technology, China. His research interests are broadly in model compression, knowledge distillation and foundation model application. 
He has published papers in ECCV. 
\end{IEEEbiography}

\begin{IEEEbiography}[{\includegraphics[width=1in,height=1.25in,clip,keepaspectratio]{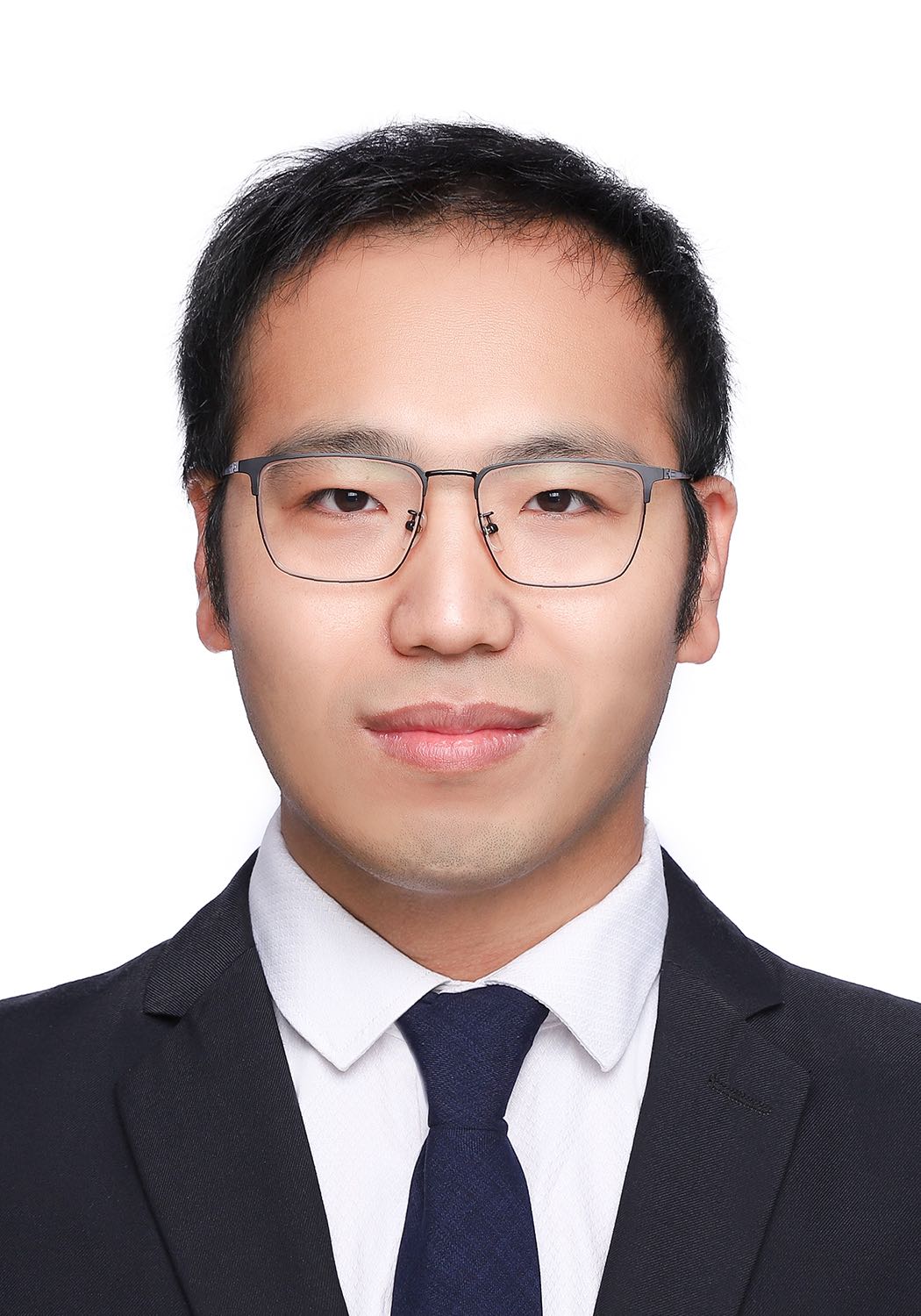}}]{Jiangchao Yao}
is an Assistant Professor of Shanghai Jiao Tong University, Shnaghai China. He received the B.S. degree in information engineering from South China University of Technology, Guangzhou, China, in 2013. He got a dual Ph.D. degree under the supervision of Ya Zhang in Shanghai Jiao Tong University and Ivor W. Tsang in University of Technology Sydney. His research interests include deep representation learning and robust machine learning.
\end{IEEEbiography}

\begin{IEEEbiography}
[{\includegraphics[width=1in,height=1.25in,clip,keepaspectratio]{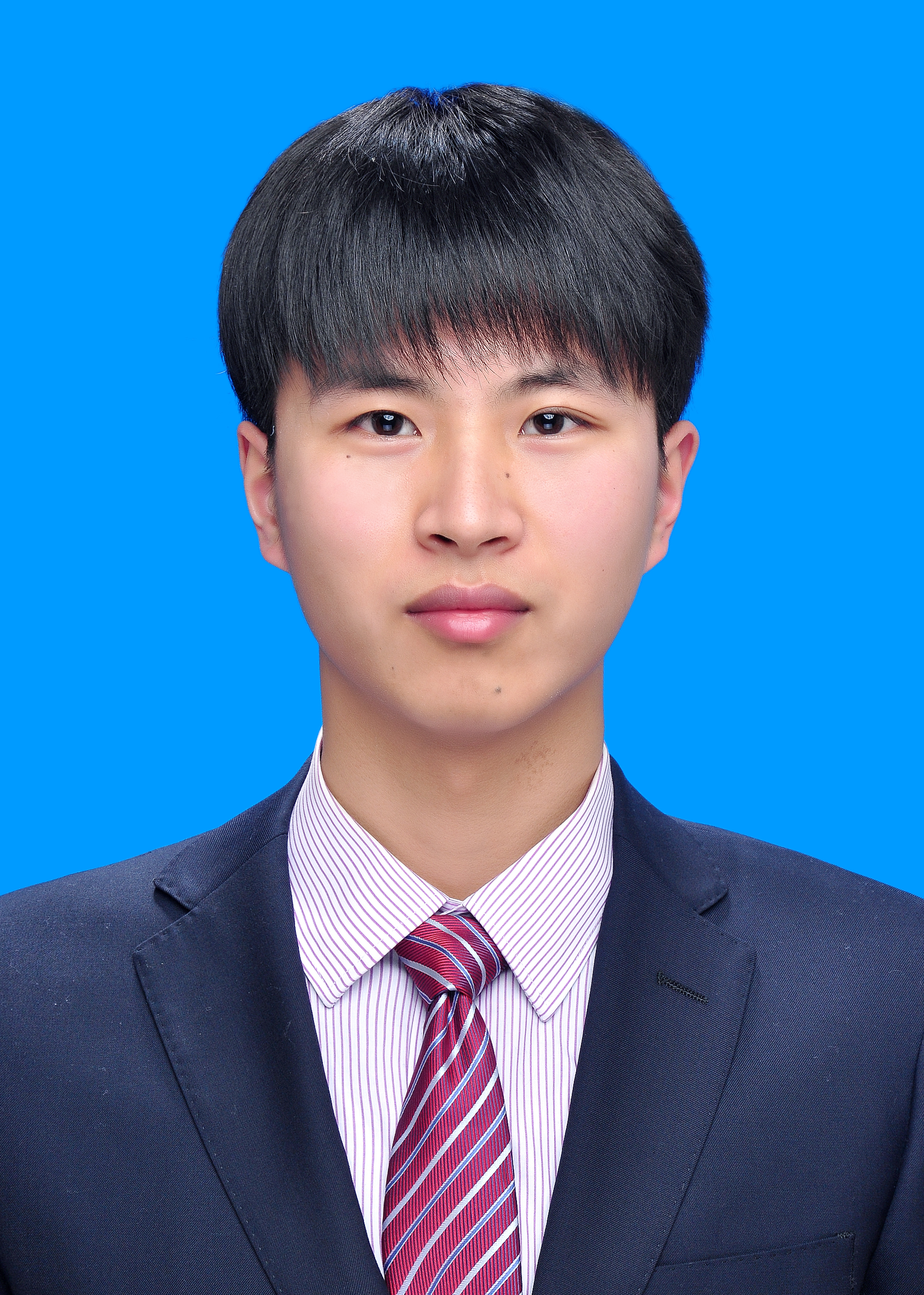}}]{Ran Luo}
is working towards a postgraduate degree in software engineering at the South China University of Technology, China. His research mainly interests in knowledge distillation, model reprogramming and semantic segmentation.
\end{IEEEbiography}

\begin{IEEEbiography}[{\includegraphics[width=1in,height=1.25in,clip,keepaspectratio]{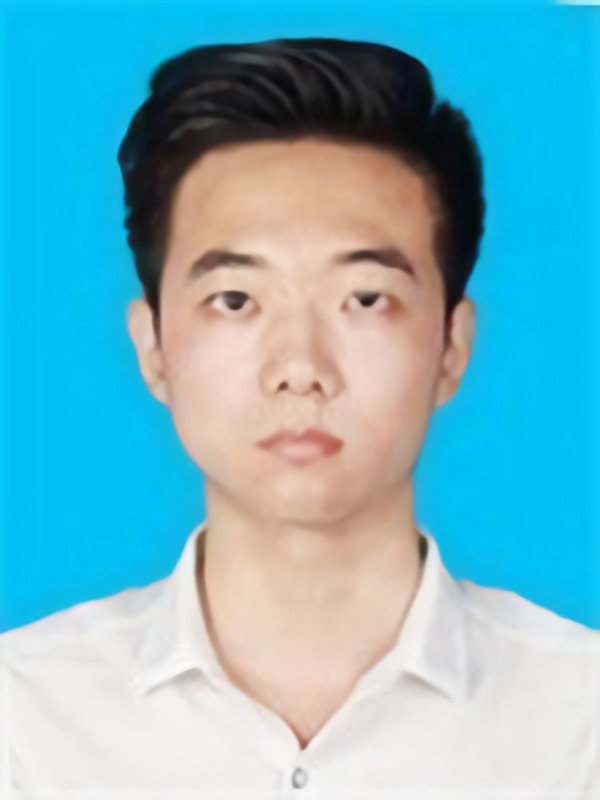}}]{Shuhai Zhang}
is currently a Ph.D. candidate at South China University of Technology, China. His research interests are broadly in machine learning and mainly focus on model compression and adversarial attack. He has published papers in Neural Networks. 
\end{IEEEbiography}

\begin{IEEEbiography}[{\includegraphics[width=1in,height=1.25in,clip,keepaspectratio]{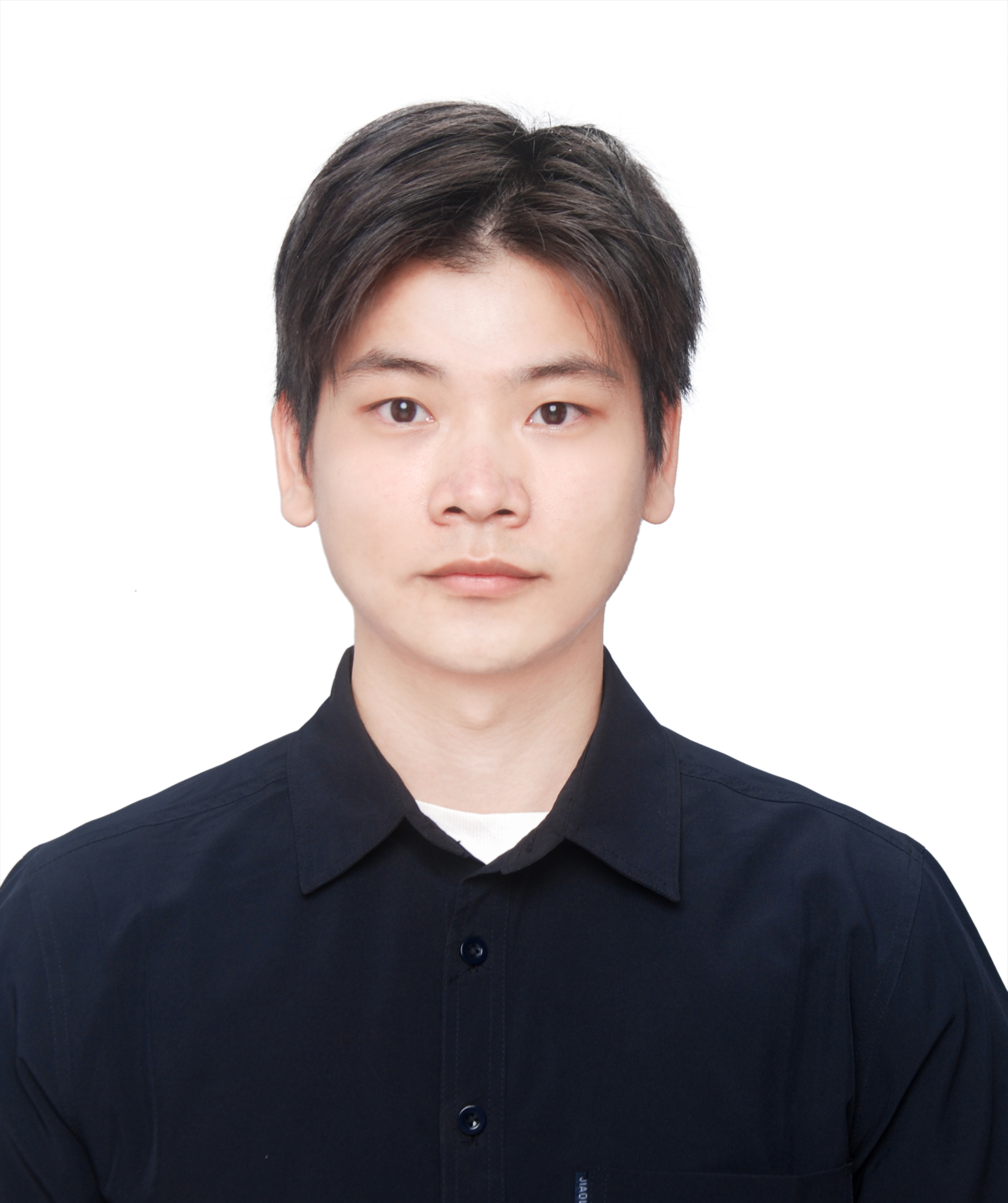}}]{Zihao Lian}
is working toward a postgraduate degree in software engineering at the South China University of Technology, China. His research interests include reinforcement learning, test-time adaptaion, point cloud deep learning, and model compression.
\end{IEEEbiography}

\begin{IEEEbiography}[{\includegraphics[width=1in,height=1.25in,clip,keepaspectratio]{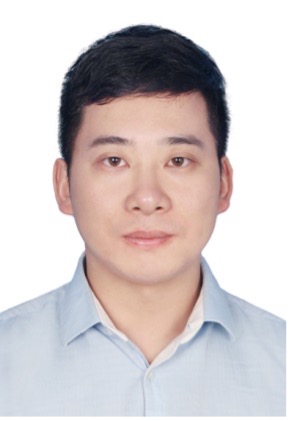}}]{Mingkui Tan}
is currently a professor with the School of Software Engineering at South China University of Technology. He received his Bachelor Degree in Environmental Science and Engineering in 2006 and Master degree in Control Science and Engineering in 2009, both from Hunan University in Changsha, China. He received the Ph.D. degree in Computer Science from Nanyang Technological University, Singapore, in 2014. From 2014-2016, he worked as a Senior Research Associate on computer vision in the School of Computer Science, University of Adelaide, Australia. His research interests include machine learning, sparse analysis, deep learning and large-scale optimization.
\end{IEEEbiography}

\begin{IEEEbiography}[{\includegraphics[width=1in,height=1.25in,clip,keepaspectratio]{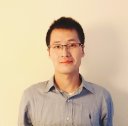}}]{Bo Han}
is currently an Assistant Professor in Machine Learning and a Director of Trustworthy Machine Learning and Reasoning Group at Hong Kong Baptist University, and a BAIHO Visiting Scientist at RIKEN Center for Advanced Intelligence Project (RIKEN AIP). He was a Visiting Faculty Researcher at Microsoft Research (2022) and a Postdoc Fellow at RIKEN AIP (2019-2020). He received his Ph.D. degree in Computer Science from University of Technology Sydney (2015-2019). During 2018-2019, he was a Research Intern with the AI Residency Program at RIKEN AIP. He has co-authored a machine learning monograph, including Machine Learning with Noisy Labels (MIT Press). He has served as area chairs of NeurIPS, ICML, ICLR and UAI, and senior program committees of KDD, AAAI and IJCAI. He has also served as action (associate) editors of Transactions on Machine Learning Research and IEEE Transactions on Neural Networks and Learning Systems, and editorial board members of Journal of Machine Learning Research and Machine Learning Journal.
\end{IEEEbiography}

\begin{IEEEbiography}[{\includegraphics[width=1in,height=1.25in,clip,keepaspectratio]{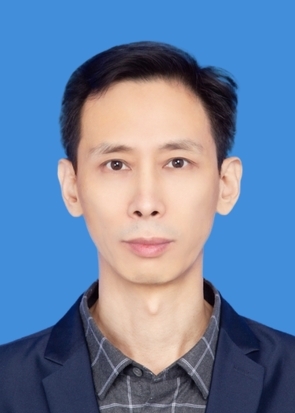}}]{Yaowei Wang}
is currently a researcher at Peng Cheng Laboratory, Shenzhen, China. 
He received the Ph.D. degree in computer science from Graduate University, Chinese Academy of Sciences, in 2005. He was an Assistant Professor with the School of Information and Electronics, Beijing Institute of Technology, and also was a Guest Assistant Professor with the National Engineering Laboratory for Video Technology, Peking University, China. He has been the author or co-author of over 50 refereed journals and conference papers. His research interests include machine learning and multimedia content analysis and understanding.
\end{IEEEbiography}







\end{document}